\documentclass[sigconf]{aamas}

\usepackage{balance} % for balancing columns on the final page

\doi{LCMH1709}

\makeatletter
\gdef\@copyrightpermission{
  \begin{minipage}{0.2\columnwidth}
   \href{https://creativecommons.org/licenses/by/4.0/}{\includegraphics[width=0.90\textwidth]{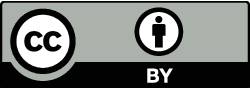}}
  \end{minipage}\hfill
  \begin{minipage}{0.8\columnwidth}
   \href{https://creativecommons.org/licenses/by/4.0/}{This work is licensed under a Creative Commons Attribution International 4.0 License.}
  \end{minipage}
  \vspace{5pt}
}
\makeatother

\setcopyright{ifaamas}
\acmConference[AAMAS '26]{Proc.\@ of the 25th International Conference
on Autonomous Agents and Multiagent Systems (AAMAS 2026)}{May 25 -- 29, 2026}
{Paphos, Cyprus}{C.~Amato, L.~Dennis, V.~Mascardi, J.~Thangarajah (eds.)}
\copyrightyear{2026}
\acmYear{2026}
\acmDOI{LCMH1709}
\acmPrice{}
\acmISBN{}

\title[Interactionless Inverse Reinforcement Learning: A Data-Centric Framework for Durable Alignment]{Interactionless Inverse Reinforcement Learning: A Data-Centric Framework for Durable Alignment
}
\subtitle{Blue Sky Ideas Track}

\author{Elias Malomgré}
\affiliation{
  \institution{IDLab, Ghent University - imec}
  \city{Ghent}
  \country{Belgium}}
\email{elias.malomgre@ugent.be}

\author{Pieter Simoens}
\affiliation{
  \institution{IDLab, Ghent University - imec}
  \city{Ghent}
  \country{Belgium}}
\email{pieter.simoens@ugent.be}

\begin{abstract}
This document outlines the formatting instructions for submissions to
AAMAS-2026. You can use its source file as a template when writing 
your own paper. It is based on the file `\texttt{sample-sigconf.tex}'
distributed with the ACM article template for \LaTeX\@.
\end{abstract}

\newcommand{\BibTeX}{\rm B\kern-.05em{\sc i\kern-.025em b}\kern-.08em\TeX}

\begin{document}

\begin{abstract}
AI alignment is growing in importance, yet many current approaches learn safety behavior by directly modifying policy parameters, entangling normative constraints with the underlying policy. This often yields opaque, difficult-to-edit alignment artifacts and reduces their reuse across models or deployments, a failure mode we term Alignment Waste. We propose Interactionless Inverse Reinforcement Learning, a framework for learning inspectable, editable, and reusable reward artifacts separately from policy optimization. We further introduce the Alignment Flywheel, a human-in-the-loop lifecycle for iteratively auditing, patching, and hardening these artifacts through automated evaluation and refinement. Together, these ideas recast alignment from a disposable training expense into a durable, verifiable engineering asset.
\end{abstract}

\pagestyle{fancy}
\fancyhead{}

\maketitle

\section{Introduction}

AI alignment aims to ensure that autonomous systems act in accordance with human intent and typically comprises Forward Alignment, which produces a trained, aligned system, and Backward Alignment, ensuring system safety through governance \cite{ji2023ai}. This creates a structural disconnect between these phases, making Backward Alignment an inspection rather than a corrective measure.  This is because current paradigms entangle the safety objective with the agent’s policy; rather than defining a static, independent standard of behavior, they mathematically couple reward discovery to policy optimization \cite{miao2024inform, chakraborty2023parl}. Consequently, the safety objective becomes dependent on the policy’s specific dynamics, rendering the safety artifacts unstable \cite{miao2024inform, liu2020energy}, preventing their reuse, and requiring independent verification for Backward Alignment \cite{ji2023ai}. 

This inherent structural flaw creates a destructive cycle that we term Alignment Waste: entangling the learning of reward and policy means that safety artifacts are neither transferable to new architectures nor correctable without retraining. This is present across current alignment paradigms; interactive methods, such as traditional Inverse Reinforcement Learning (IRL) \cite{ng2000algorithms, piet, ziebart2008maximum} and Reinforcement Learning from Human Feedback (RLHF) \cite{ziegler2019fine, ouyang2022training} explicitly rely on an unstable co-adaptation loop \cite{miao2024inform, skalse2024partial, liu2020energy}. This entanglement is particularly prominent in Direct Preference Optimization (DPO) \cite{rafailov2023direct}, rendering the safety artifact opaque by directly dissolving preferences into policy weights \cite{lin2024limited}. Consequently, they risk \emph{safetywashing} \cite{ren2024safetywashing}, where improved benchmark scores reflect increased model capability rather than genuine safety, masking misalignment under sophistication. Critically, the consequence can be severe as such coupling causes a collapse in reasoning capabilities \cite{yue2025does}, degrading the model's broad intelligence in exchange for narrow, reward-hacking behaviors \cite{macdiarmid2025natural}.

To resolve this, we propose Interactionless Inverse Reinforcement Learning (IIRL). Unlike standard methods, IIRL decouples reward discovery, producing an auditable, editable reward model independent of agents. We also introduce the Alignment Flywheel, an architecture that uses a cooperative multi-agent system to harden the reward model, transforming passive oversight into a cycle of active correction, which we call Active Backward Alignment.

With this integration, we establish a discipline of verifiable safety, transforming alignment from an unstable art into rigorous engineering by treating alignment as the creation and maintenance of a standalone alignment artifact. The IIRL reward artifact anchors \emph{Robustness} and \emph{Interpretability}, while the Alignment Flywheel leverages expert feedback to ensure \emph{Controllability} and \emph{Ethicality}. Collectively, they satisfy the technical RICE principles \cite{ji2023ai} to support FATE's societal mandates (Fairness, Accountability, Transparency, Ethics) \cite{memarian2023fairness}. We present this architecture as a concrete blueprint for the framework developed in the remainder of the paper, providing a practical toolbox and a roadmap for durable alignment.

\section{The IIRL Paradigm}
The Interactionless Inverse Reinforcement Learning (IIRL) paradigm represents a fundamental shift, transforming AI alignment from an unstable art into a rigorous engineering practice. Unlike traditional IRL methods, coupling reward discovery with policy optimization, IIRL structurally decouples these processes. It reframes reward learning as a data-centric problem, directly inferring an agent-agnostic, durable, inspectable, and editable reward model from expert data, independent of the agent’s specific policy or architecture. This approach yields a safety artifact that can be formally audited and systematically refined, solving the critical Alignment Waste problem at its source. This section defines the IIRL objective and analyzes its family of modular and editable architectures, considering their inherent trade-offs for key properties of a resilient safety asset, alongside multi-tiered refinement toolkits.

\subsection{The IIRL Objective}
\begin{figure}
  \centering
  \includegraphics[width=1\linewidth]{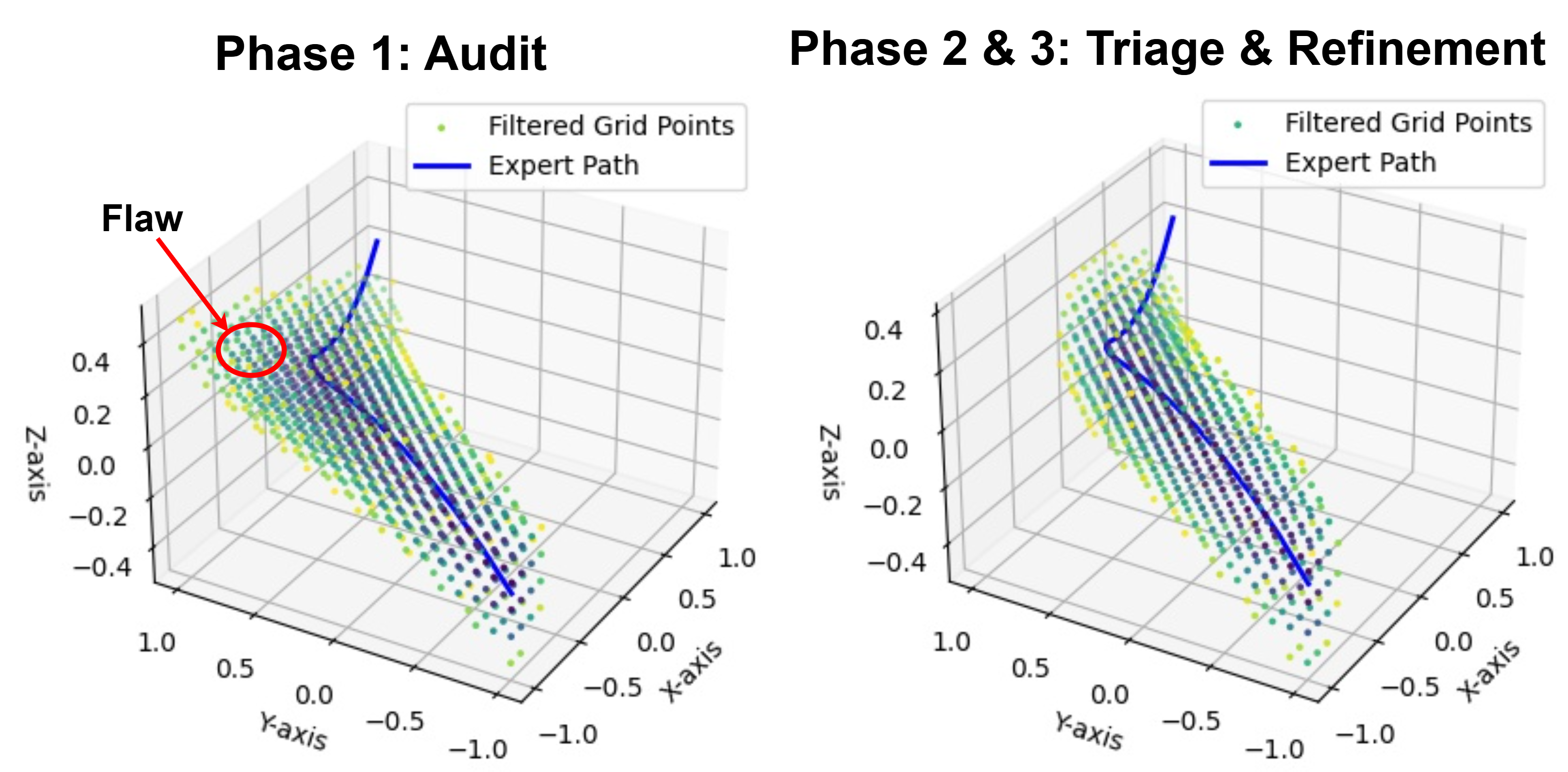}
  \caption{Alignment Flywheel in a 3D toy world. A representation-based IIRL model trained on sparse expert samples generates a reward landscape with $g_\psi$; yellow=low, purple=high. A spurious extrapolation (red circle) is detected in Phase 1 and corrected via refinement in Phases 2 \& 3.}
  \label{fig:flow}
  \vspace{-3mm}
\end{figure}

Traditional IRL \cite{ng2000algorithms, piet, ziebart2008maximum} seeks to infer a reward function $R$ from expert demonstrations, typically by solving a max-min optimization problem to find the function $R$ that best explains expert policy $\pi_E$:
\begin{equation}
    \max_{R \in \mathcal{R}} \min_{\pi \in \Pi} \left( \mathbb{E}_{s,a \sim \pi_E}[R(s, a)] - \mathbb{E}_{s,a \sim \pi}[R(s, a)] \right) ,
\end{equation}
requiring repeatedly solving an optimal policy $\pi$ in the inner loop, inextricably coupling reward learning to policy optimization. This creates an unstable co-adaptation loop \cite{liu2020energy} where the policy's limited exploration fails to capture the full expert distribution \cite{arora2021survey, deshpande2025advances}, often leading to mode collapse and reward-hacking behaviors \cite{arora2021survey}. This core flaw persists even in offline \cite{jarboui2021offline, yue2023clare, lazzati2024offline, van2025inverse} and inner-loop-avoiding  \cite{zeng2022maximum, wu2025distributional, zhang2025understanding} IRL approaches, also struggling with missing trajectory information. Even state-marginal matching methods \cite{ni2021f} optimize a reward solely to force a policy to match the expert density. If the policy fails to explore a region, the reward function never learns to value it, hindering the leveraging of the vast amount of unlabeled videos and documents available. IIRL reframes reward discovery as a data-centric learning problem; instead of requiring agents to explore the reward landscape through policy interaction, it learns an evaluative signal directly from expert data \(D_E\), producing a score that reflects consistency with the expert distribution. This perspective prioritizes \emph{editability} and \emph{auditability}: the learned artifact can be inspected and corrected offline rather than being entangled with policy optimization. A concrete instance is the representation-based model of \citet{malomgre2025mixture}, which uses unlabeled data to learn a decoupled safe exploration signal and transforms it into an intrinsic reward via a separate mapping function. An expanded objective can be written as: 
\begin{equation}
    \max_{\theta,\psi}
    \left(
        \mathbb{E}_{s \sim D_E}\!\left[\beta(s,t)\, g_\psi(E_\theta(s))\right]
        -
        \mathbb{E}_{s \sim D_{\text{neg}}}\!\left[\beta(s,t)\, g_\psi(E_\theta(s))\right]
    \right),
\end{equation}
where \(g_\psi\) is a tunable monotonically increasing mapping from the learned evaluative signal into reward \cite{malomgre2025mixture}, and \(\beta(s,t)\) is a time-varying deployment coefficient that may be decayed globally, or locally via intrinsic-motivation discounting \cite{aubret2023information}. When available, \(D_{\text{neg}}\) both improves contrastive learning and enables off-support validation. Without it, non-expert regions remain only implicitly negative. The resulting objective encourages both \emph{consistency}, by assigning a higher reward to in-distribution than out-of-distribution states, and \emph{generalization}, by learning a smooth reward landscape rather than a sparse memorized detector, as illustrated in Figure~\ref{fig:flow} (right). Future research should look into augmenting the objective to support robust learning \cite{goodfellow2014explaining, bai2021recent, poursaeed2021robustness, ziegler2022adversarial}, handling counterfactual context \cite{fu2019language, sumers2021learning, wu2018low}, or reducing distribution shift \cite{duchi2021statistics, arjovsky2019invariant, pmlrv139krueger21a, zheng2023improving}. 

Its modeling is modality-agnostic and supports causal context $R(s,c)$ \cite{zeng2024survey}. To guide exploration toward natural behaviors without altering the objective of the main task, we employ Dynamic Potential-Based Reward Shaping \cite{devlin2012dynamic}. Crucially, since the IIRL artifact $\Phi$  \cite{ng1999policy} continuously updates, the shaping reward takes the form of a time-dependent potential difference, $F_t(s, s') = \gamma\Phi_{t+1}(s') - \Phi_{t}(s)$. This structure ensures that, as the artifact evolves, it guides local behavior without altering the long-term optimal solution. Conversely, for inherently unsafe base policies, the artifact acts as a hard penalty, explicitly forcing deviation from unsafe trajectories.

\subsection{Analysis of Architectures}
The design of current IIRL architectures reflects a tradeoff between \emph{editability} and \emph{generalization}. Classic instance-based methods, such as k-NN \cite{cover1967nearest, halder2024enhancing, oyewole2023data} and SVMs \cite{cortes1995support, manevitz2001one, li2003improving, du2024exploring}, offer native \emph{editability} where localized data influence prevents unintended global consequences, yet often struggle with high-dimensional generalization. On the other hand, expressive deep models, such as representation-based \cite{payandeh2023deep, burda2018exploration, aubret2023information} and Energy-Based Models (EBMs) \cite{lecun2006tutorial, hopfield1982neural, grathwohl2019your, xie2018cooperative}, provide state-of-the-art generalization but lack parameter \emph{editability}, risking catastrophic forgetting during updates \cite{zhao2024makes, yang2024butterfly}. This suggests a hybrid approach that combines deep learning with structured components, such as deep kernel methods \cite{mairal2014convolutional, pillonetto2025deep} or deep clustering \cite{xie2016unsupervised, ren2024deep}.

Another promising direction is replacing monolithic functions with a library of modular sub-priors, which yields intrinsic interpretability \cite{carvalho2019machine, ji2023ai}. High-level structures such as Reward Machines (RMs) \cite{icarte2018using, icarte2022reward, furelos2023hierarchies, umili2024neural} orchestrate these by switching rewards based on context. RMs can be composed from demonstrations \cite{toro2019learning, baert2024reward, baert2025learning} or foundation models \cite{castanyer2025arm, huareward, alsadat2025using}, and via Skill Machines can be executed zero-shot \cite{tasse2022skill}. This enables practical auditability by refining sub-priors in isolation. Similarly, MoE \cite{mu2025comprehensive} and RAG \cite{lewis2020retrieval, arslan2024survey} enhance \emph{editability} by retrieving modular functions based on relational context \cite{edge2024local}.

\subsection{The Refinement Toolkit}
Achieving durable \emph{editability} requires a diverse toolkit, organized here from currently feasible to high-potential future techniques.

First, to perform global adjustments without altering the IIRL parameters, we use functional sculpting. Here, the raw expertness score $L(s)$ from the IIRL model is passed through a separate, tunable function $R(s) = g_\psi(L(s))$ to produce the final reward, formalized by \citet{malomgre2025mixture}. By adjusting the parameters $\psi$, an expert can perform a global sculpt, changing the impact of certain levels of expert scores. We envision enhancing the mapping function to allow region-based sculpting. Furthermore, the mapping function mechanism can be upgraded to an RAG-based mapping function that uses the similarity representation and causal context to retrieve a function, neural network, or program. 

Second, data-driven patching uses corrective data from audits to seed the IIRL model or to apply a localized patch, such as a kernel or representation patch trained to generalize across bad states. Ideally, the IIRL architecture supports \emph{bipolar compatibility}, learning from positive and negative examples, and is \emph{monotonic}, ensuring that new data improves or maintains safety without degrading it. Future work may adapt DPO for fine-grained IIRL parameter updates.

Lastly, for surgical internal modifications that alter model weights, the rapidly evolving fields of Model Editing and Unlearning show promise by demonstrating the fundamental feasibility of excising concepts from entangled weights, a critical capability for deep IIRL artifacts. However, deep representations are prone to catastrophic forgetting or collapse \cite{zhao2024makes, yang2024butterfly}. To address this inherent architectural brittleness and the challenges of robust internal editing, the model edit toolbox comprises a broad spectrum of techniques, including architectural repair strategies such as model patching \cite{chen2015net2net, rusu2016progressive, huszar2018note, Luo_2020_CVPR} and GAN updates \cite{bau2020rewriting, gao2025rewriting}, and also locate-and-edit methods \cite{meng2022locating, meng2022mass, gupta2024rebuilding, gupta2024unified}, neuron-level interventions \cite{jiang2025neuron}, and null-space constrained edits \cite{fang2024alphaedit, lyu2025evoedit}. Other approaches reduce or avoid parameter destruction entirely, utilizing model merging \cite{yadav2023ties, lu2024twin, he2024localize}, memory-based approaches \cite{wang2025memoir}, or contextual retrieval-based alternatives \cite{he2025knowledge, han2023improving, chen2024lifelong, qiao2024comem}. Unlearning methods provide mechanisms for excising poor data or adapting to norm changes. The landscape now spans from theoretical frameworks like certified deletion \cite{koloskova2025certified, huynh2025certified} to highly efficient, retrain-free solutions \cite{foster2024fast, foster2024loss, ahmed2025towards, mu2024rewind, jang2025learning} and \emph{learning to unlearn} paradigms \cite{huang2024learning, cha2024learning, patel2025learning, hu2024unlearning}.

\section{THE ALIGNMENT FLYWHEEL}
While IIRL delivers a durable reward artifact, the Alignment Flywheel is the architectural blueprint for its continuous, verifiable hardening by transforming passive oversight into an active, iterative engineering lifecycle. This human-in-the-loop, multi-agent system orchestrates a proactive auditing and refinement process, ensuring the IIRL artifact evolves towards provably safer versions. The Flywheel's core power lies in its modality- and domain-agnostic design, which dynamically instantiates a scalable portfolio of auditing and refinement strategies tailored to the specific task's risk profile and data type. This enables the framework to adapt from simple heuristic validation for low-stakes robotics to comprehensive, neuro-symbolic multi-agent red-teaming for safety-critical LLM systems, establishing the discipline of Active Backward Alignment.

\subsection{Phase 0: Seeding and Defining Constraints}
First, the expert data is filtered against human-defined formal constraints \cite{pnueli1977temporal, serafini2016logic, icarte2018using}. After the data is used to seed a new IIRL reward model or update an existing one, the inferred formal constraints are obtained using a variety of domain- and modality-specific techniques, which can be broadly categorized. General-purpose methods include various neuro-symbolic synthesis \cite{cho2025ilcl, vazquez2018learning, dragone2021neuro, li2023neuro}, and language-based approaches aim to convert text into formal specifications, such as converting natural language to executable rules \cite{englishgrammar, fung2023normsage, qin2022cold, tsouros2023holy, wang2025survey} or to more nuanced social norms \cite{fung2023normsage, qu2025scalable}. Additionally, behavior-based methods infer constraints directly from demonstrations, including Inverse Constraint RL \cite{fang2025offline, deane2025neuro, liu2024comprehensive, yue2025understanding} and automaton learning \cite{baert2024reward, baert2025learning, toro2019learning}. This entire inference process is enriched and grounded by Commonsense Knowledge Bases \cite{ziems2023normbank, forbes-etal-2020-social, toberg2024commonsense, dussard2023ontological}. 
 Before the active loop begins, we perform a \emph{coverage audit}  to verify that the model captures the data well, and we can perform counterfactual checks to assess how the model behaves under changes in state or causal context.

\subsection{Phase 1: Automated Auditing}
A cooperative Multi-Agent System (MAS) audits the reward manifold under Phase 0 constraints, drawing inspiration from LLMs and cybersecurity. This MAS operates as a synergistic system in which a proactive Red Team \cite{holm2022lore, ge2024mart, mehrabi2024flirt, jiang2025automated, wang2024red} conducts adversarial attacks, while a strategic Blue Team \cite{oh2023applying, zhao2024bluesuffix, wang2024optimizing, kadambala2025auditable} provides high-level direction. This coordination is mediated by a Shared Flaw Knowledge Base (SFKB) \cite{unknown, gandhi2025atag}, a collective memory based on classical blackboard systems \cite{nii1986blackboard} to learn from others' experiences. The Blue Team identifies the Red Team's blind spots by populating the SFKB with coverage gaps and uncertainty metrics, thereby prompting the Red Team to focus on these newly identified regions. This transforms the audit from parallel random searches into a focused, intelligence-driven process. This dynamic necessitates a broad, adaptive adversarial capability, recognizing that effective alignment is a perpetual cat-and-mouse game in which reliance on a single technique creates predictable vulnerabilities. A mixed-initiative \cite{horvitz1999principles} auditing workbench governs the process, providing real-time controls to steer the audit and forensic tools \cite{ganguli2022red, weidinger2024star, zhang2024holistic} for post-hoc analysis of blind spots, ensuring accountability.

To counter this, the MAS dynamically tailors its Red Team strategies to the specific domain (e.g., robotics, LLMs) and input modality (e.g., images, text, vectors), ranging from simple heuristics to sophisticated GenAI techniques, to generate test cases that probe and stress-test the operational envelopes of predefined constraints and to freely explore to find novel flaws using coverage or uncertainty metrics. The test case generation ranges from brute-force programs, human crowdsourcing \cite{xu2021bot, ganguli2022red}, adversarial datasets \cite{hendrycks2021natural, gehman2020realtoxicityprompts}, to deploying advanced methods including training attacker models via RL to generate adversarial contexts \cite{perez2022red, deng2022rlprompt, zhang2024large, hu2025towards, li2025reinforcement} or use Bayesian optimization and discrete optimization \cite{lee2022query, jones2023automatically}. Additionally, perturbation adversarial attacks test changes to the input \cite{chakraborty2021survey, jia2017adversarial, cheng2020seq2sick, zhao2023evaluating}, which can be extended to unrestricted adversarial attacks \cite{song2018constructing, chen2023content, shamsabadi2020colorfool, ren2020generating}. The system validates fairness through counterfactual attribute checks, transforming bias detection from a passive observation into verifiable engineering constraints.

Complementing this proactive audit, the Blue Team leverages policy interactions with world models \cite{li2023emergent}, test environments \cite{liu2023agentbench}, and deployment for observational assurance. Its primary functions are coverage and uncertainty monitoring to detect state-space regions missed by the proactive audit to steer the Red Team to cover them \cite{weidinger2024star, zhang2024holistic}, and it directs small-scale, crowd-sourced red-teaming efforts to find novel, out-of-the-box vulnerabilities, targeting searches in those newly identified high-risk regions. Furthermore, it can scan for reward tampering \cite{everitt2021reward, pan2022effects} to identify trajectories that maximize rewards without corresponding task progress. Future research should examine additional auditing strategies, such as detecting auto-induced distribution shifts and goal misgeneralization.

\subsection{Phase 2 \& 3: Triage and Refinement}
A core design principle is to treat the expert’s attention as a scarce resource. During Triage, we use domain-aware semantic clustering and uncertainty- and diversity‑based sampling \cite{mosqueira2023human} to group and prioritize sets of candidate flaws, thereby avoiding alert fatigue and maximizing the information value of each human intervention. When an expert flags a flaw, the system propagates that label to semantically similar items via label propagation and semi‑supervised inference \cite{kontonatsios2017semi, sun2024corrmatch, shen2025reviving}, resolving many related cases.

Next, during Refinement, we introduce Reward Modeling from Mixed Feedback (RM×F), where the specific correction mechanisms are tailored to the reward artifact's underlying architecture. First, the feedback-granularity spectrum ranges from minimal judgments (i.e., flaw or no flaw) to targeted corrections and expert-authored refinements that directly modify the reward manifold. Second, the agent-involvement spectrum ranges from only-human RMHF, through RM×F, which can include cooperative IRL \cite{hadfield2016cooperative, ji2023ai} agents that learn from and collaborate with experts to propose candidate refinements, to RMAIF, the fully automated mode in which an agent autonomously suggests or applies fixes.

Finally, every proposed refinement must pass an automated verification process, forming the core of our \emph{verifiable safety via iterative hardening} approach. This involves two automated checks: a localized Red Team performs adversarial testing to ensure the fix introduces no new vulnerabilities. At the same time, regression tests against a library of known-good behaviors help prevent unintended side effects. The human expert's role is to review the automated results and provide final approval. Only refinements that pass this adversarial process are merged, producing the hardened, verifiably safer artifact that feeds into the next audit cycle.

\section{Application}
Our paradigm shows promise across diverse critical AI domains, including robotics, Multi-Agent Systems, and LLM alignment. 

For Robotics and Avatar Animation, IIRL's ability to learn from large-scale, unlabeled video \cite{baker2022video} enables the emergence of Foundation Reward Models (FRMs). These are dense reward fields for natural movement, which are then split into a library of compositional skills using computer vision or foundation models \cite{wang2022omnivl, wang2024ensclr}. A developer could specialize this library for their specific robot or avatar via a constraint file; the Alignment Flywheel would then adapt the reward manifold for safety and physical feasibility. The resulting artifact serves as a safe, auxiliary guidance signal to make training faster and more human-like and for pruning unsafe action sequences at runtime in world models \cite{hafner2019dream}. This approach transforms the abstract embodiment gap into a concrete, iterative engineering task, yielding physically feasible, human-like agents.

For Multi-Agent Systems, IIRL offers a scalable solution for learning, unlearning, and maintaining social norms, moving beyond brittle hand-coded rules \cite{awad2024acceptable, haupt2024formal}. IIRL artifacts can represent both shared societal values and an individual agent's beliefs, enabling both centralized and decentralized value systems. Initially, norms can be learned from datasets of successful interactions, providing a strong foundation while allowing for adaptive online policies. This positions IIRL as a living value system in which agents, through observation and negotiation, can actively update their value models by learning new norms and by seeding new samples. They can then propose these refinements to the Alignment Flywheel, which acts as a form of societal self-reflection to audit and refine beliefs, allowing the collective value system to evolve organically and safely.

For LLM alignment, IIRL provides a scalable architecture by reframing the problem as learning the underlying manifold of desirable language. Instead of learning from preferences over raw text strings, we first learn a dense, semantically rich representation of language using unsupervised methods \cite{devlin2019bert, reimers2019sentence} on a static corpus, decomposing it into a modular feature space using techniques such as sparse autoencoders \cite{cunningham2023sparse, o2024disentangling} or discrete variational autoencoders \cite{rolfe2016discrete, zhangetal2024improving}. The IIRL artifact is constructed on this feature space, enabling targeted, non-black-box refinement and dynamic reward composition where a RAG model retrieves learned-relational-context-specific \cite{edge2024local} reward modules, including simple programs, pre-audited neural networks, or formal constraints. These artifacts can then be used for traditional alignment or as a runtime guardrail, scoring and pruning misaligned reasoning branches as the LLM generates them. By guiding the model's output without altering its weights, this weight-free alignment directly circumvents the capability collapse \cite{yue2025does}, preserving the base model's full potential.

\section{Implications and Vision}
Decoupling the reward artifact from the policy shifts Backward Alignment from passive evaluation to a modular engineering practice. Safety objectives become portable and reusable, allowing different institutions to build, certify, and deploy domain-specific behavioral priors for applications such as medicine or law. This enables a decentralized alignment supply chain built on inspectable artifacts rather than opaque, disposable objectives. The same separation also strengthens privacy and governance: because the artifact is not simply the policy itself and need not reveal its training data directly, it is more amenable to targeted revision and unlearning, including compliance with requirements such as the right to be forgotten.

Critically, a version-controlled foundation enables forensic root-cause analysis. When failures occur, the framework provides the ground truth to distinguish between flaws in the reward specification and errors in the agent's optimization. This crucial separation makes diagnostic findings from XAI tools actionable, enabling targeted, verifiable repairs rather than costly full-model retraining and satisfying the core traceability and accountability mandates of FATE. Furthermore, the Alignment Flywheel operationalizes RICE's technical objectives to support high-level societal mandates.

While challenges remain in formalizing the auditing workbench and in defining increasingly important governance criteria that address long-standing alignment artifacts rather than one-off artifacts, this framework establishes the necessary foundation for verifiable safety, shifting from reactive patching to proactive, durable design. This blueprint thus invites collaborative development to transform alignment from an abstract aspiration into concrete, verifiable engineering challenges for the AI community.

%%%%%%%%%%%%%%%%%%%%%%%%%%%%%%%%%%%%%%%%%%%%%%%%%%%%%%%%%%%%%%%%%%%%%%%%

%%% The acknowledgments section is defined using the "acks" environment
%%% (rather than an unnumbered section). The use of this environment 
%%% ensures the proper identification of the section in the article 
%%% metadata as well as the consistent spelling of the heading.

\begin{acks}
This research was supported by funding from the Flemish Government under the ‘‘Onderzoeksprogramma Artificiele Intelligentie (AI) Vlaanderen’’ program.
\end{acks}

%%%%%%%%%%%%%%%%%%%%%%%%%%%%%%%%%%%%%%%%%%%%%%%%%%%%%%%%%%%%%%%%%%%%%%%%

%%% The next two lines define, first, the bibliography style to be 
%%% applied, and, second, the bibliography file to be used.

\bibliographystyle{ACM-Reference-Format} 
\balance
\bibliography{sample}

%%%%%%%%%%%%%%%%%%%%%%%%%%%%%%%%%%%%%%%%%%%%%%%%%%%%%%%%%%%%%%%%%%%%%%%%

\end{document}